\RequirePackage[svgnames,table]{xcolor}
\documentclass[11pt,letterpaper]{yalearxiv}

\usepackage{microtype}
\usepackage{natbib}
\setcitestyle{square}

\makeatletter
\def\munderbar#1{\underline{\sbox\tw@{$#1$}\dp\tw@\z@\box\tw@}}
\makeatother

\AddToHook{cmd/appendix/before}{%
  \setcounter{axiom}{0}%
}

\newcommand{\be}{\begin{equation}}
\newcommand{\ee}{\end{equation}}
\newcommand{\bea}{\begin{equation*}\begin{aligned}}
\newcommand{\eea}{\end{aligned}\end{equation*}}

\title{LifeFuse-Mem: Lifecycle-Aware State Fusion Against Temporary Overwriting for Long-Term Memory}

\author{
Hanyu Zhao\textsuperscript{1,*},
Yuqian Feng\textsuperscript{1,2,3,*},
Zhenyu Song\textsuperscript{4,*},
Yuanchao Cheng\textsuperscript{1},
Yance Jiao\textsuperscript{1},
Tengfei Pan\textsuperscript{1,$\dagger$},
Li Du\textsuperscript{1,$\dagger$}\\
\textsuperscript{1}Beijing Academy of Artificial Intelligence\\
\textsuperscript{2}University of Chinese Academy of Sciences\\
\textsuperscript{3}Institute of Software, Chinese Academy of Sciences\\
\textsuperscript{4}National University of Defense Technology\\
}

\footnotetext[1]{$^{*}$These authors contributed equally to this work. Emails: hyzhao@baai.ac.cn}
\footnotetext[2]{$^{\dagger}$Corresponding author.}

\hypersetup{colorlinks=true, linkcolor=blue!50!black, citecolor=blue!50!black,
            urlcolor=blue!50!black}
            
\begin{document}

\begin{abstract}
\vspace{1em}
Long-running LLM agents require memory mechanisms that maintain coherent internal states across interactions. We study a lifecycle-labeled memory setting in which write episodes provide lifecycle metadata during training, and phase-aware readout is used during evaluation. This setting reflects the need to distinguish information that should remain influential across future interactions from information that should affect only the current context. A mismatch between these lifecycles can cause temporary information to overwrite durable knowledge, leading to behavioral drift in persistent agents. Within this setting, we introduce \textbf{LifeFuse-Mem}, a lifecycle-aware neural memory framework that separates information according to its temporal commitment. LifeFuse-Mem uses dedicated memory components and lifecycle-aware updates to allow stable and transient knowledge to evolve locally without converting temporary context into durable state. On the controlled anti-overwrite benchmark, LifeFuse-Mem improves acquisition-controlled retention and reduces temporary overwrite; on two public long-memory benchmarks, it remains broadly competitive. These results suggest that explicit lifecycle signals can help diagnose and mitigate overwrite in compact online memory.
\end{abstract}

\maketitle

\section{Introduction}

\begin{figure}[t]
\centering
\includegraphics[width=0.72\linewidth]{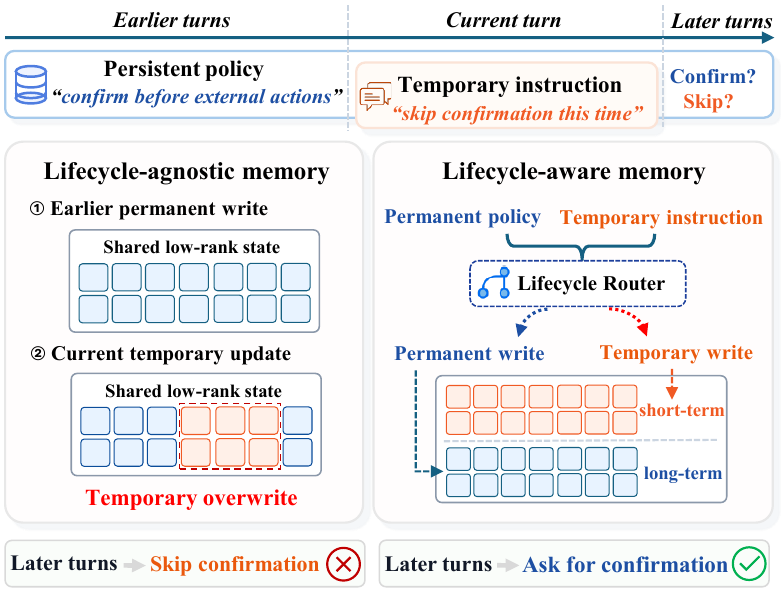}
\caption{Temporary overwrite arises when a temporary assumption is written into
the same compact state as a permanent fact. LifeFuse-Mem separates information
lifecycles at write time and uses phase-aware readout to preserve long-term
facts while still following local assumptions.}
\label{fig:intro-temporary-overwrite}
\end{figure}

Large language model (LLM)-based agents are evolving from short-lived
assistants into persistent systems that continuously interact with
users, accumulate experience, and execute long-horizon tasks~\cite{ye2026agentfold,cheng2026amemgym}. To support
such persistent interactions, agents increasingly rely on memory
mechanisms to maintain a coherent state beyond the current context
window, including user preferences, behavioral policies, and accumulated
knowledge~\cite{wu2025longmemeval,hu2026memoryagentbench}. However, unlike conventional information storage systems,
agent memory must manage information that differs fundamentally in
temporal validity: some information should persist across future
interactions, while other information should only influence the current
situation~\cite{ong2025theanine}.

Existing memory systems~\cite{xu2026mem,tan2025rmm} largely overlook this lifecycle dimension,
creating a fundamental failure mode in long-running agents. When
information with different persistence requirements is stored and
updated through the same memory pathway, temporary context can overwrite
durable knowledge. For example, an agent may maintain a persistent
policy requiring user confirmation before external actions. During a
specific debugging session, the user may temporarily instruct the agent
to skip confirmations because all operations occur in a sandbox. The
agent should adapt to this local assumption while preserving its
original policy for future tasks. Without lifecycle-aware memory,
however, the temporary instruction may be incorporated as a lasting
update, causing the agent to gradually drift from its intended behavior.
We refer to this failure as temporary overwrite. Such failures are
particularly problematic for autonomous agents because they silently
alter future decisions, degrade behavioral consistency, and undermine
user trust.

The root cause of temporary overwrite is the lack of lifecycle
disentanglement in memory representations. Existing neural memory
mechanisms typically compress interaction histories into a shared latent
state and update this state using unified dynamics~\cite{behrouz2026titans,wang2025mplus}. Consequently,
transient assumptions and persistent knowledge compete for the same
representational capacity: an update intended for short-term adaptation
can modify the latent components encoding long-term information.
Preventing temporary overwrite therefore requires more than increasing
memory capacity or improving retrieval; it requires a memory
architecture that explicitly separates information according to how long
it should influence future agent behavior.

To address this challenge, we study a controlled but diagnostic lifecycle-labeled memory setting in which write episodes specify whether incoming evidence should be treated as durable or temporary, and phase-aware readout uses known phase boundaries and query lifecycles at evaluation time. This setting is intended to isolate temporary overwrite under explicit lifecycle signals; it does not claim autonomous lifecycle discovery from unlabeled interaction histories. Our contributions are threefold: first, we define temporary overwrite as a lifecycle mismatch failure mode and evaluate it with an acquisition-controlled protocol; second, we build Hard Attribution Anti-Overwrite as a diagnostic benchmark for measuring whether permanent facts survive conflicting temporary writes; third, we propose LifeFuse-Mem, which reduces overwrite in compact online memory states when lifecycle metadata and phase structure are available.

We evaluate LifeFuse-Mem in two roles. The controlled Hard Attribution
Anti-Overwrite benchmark provides the direct anti-overwrite test, where
LifeFuse-Mem improves acquisition-controlled retention under explicit lifecycle
supervision and phase-aware readout. LoCoMo and MemoryAgentBench do not provide
overwrite phases or query lifecycle labels, so we use them only as compatibility
checks for the trained memory adapter in standard online-memory mode; in this
setting, LifeFuse-Mem remains broadly competitive, with gains on LoCoMo and
mixed category-level results on MemoryAgentBench.

\begin{figure*}[t]
\centering
\includegraphics[width=0.9\textwidth]{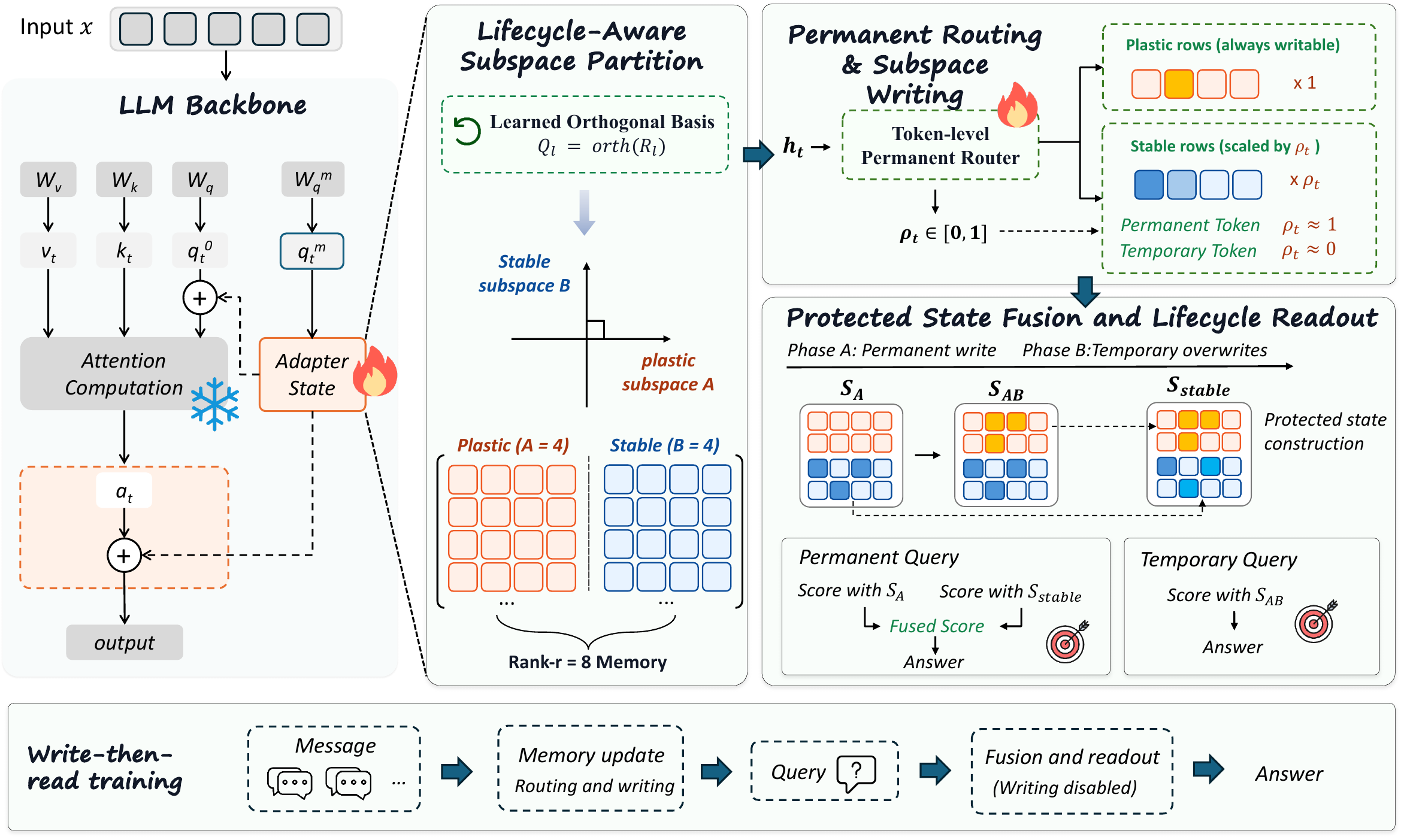}
\caption{Overview of LifeFuse-Mem. The method separates plastic and stable
subspaces, routes permanent tokens toward stable writing, and uses
phase-aware protected-state readout.}
\label{fig:method-overview}
\end{figure*}

\section{Related Work}

\subsection{Memory-Augmented Language Models}
Memory-augmented language models extend context through retrieval, replay,
compression, or neural state. Retrieval systems store passages or hidden states
and fetch them at generation time, including retrieval-augmented
pre-training~\cite{zhao2025funnelrag,jin2025longcontextrag}, retrieve-and-read
generation~\cite{gao2025trainlongcontext,cheng2025coral}, $k$NN or datastore-based
lookup~\cite{wang2025mplus,zanzotto2025memo}, and
self-reflective retrieval~\cite{tan2025rmm,ong2025theanine}. Long-context methods instead
compress, index, or manage history through compressive
memory~\cite{behrouz2026titans,he2025hmt}, landmark access~\cite{gao2025trainlongcontext,bai2025longbenchv2},
gist tokens~\cite{he2025hmt,wu2025lifbench,deng2025gist}, or tiered memory
management~\cite{kang2025memoryos,chhikara2025mem0}, though longer replay alone does not ensure
reliable use of all positions~\cite{bai2025longbenchv2,wu2025lifbench,yen2025helmet,modarressi2025nolima}. LifeFuse-Mem is closest to
parametric or adapter memories that are read during the forward pass, including
selective state-space memory~\cite{he2025hmt}, test-time-training
states~\cite{behrouz2026titans}, neural long-term memory modules~\cite{behrouz2026titans,wang2025mplus},
delta-rule associative states~\cite{lei2026deltamem}, trainable memory
layers~\cite{zanzotto2025memo}, external or latent neural memory
modules~\cite{xu2026mem,wang2025mplus}, controllable key--value
memory~\cite{liao2026sparta}, and decoupled side-networks~\cite{hu2025hiagent,wang2025workflowmemory}.
In particular, $\delta$-Mem stores interaction evidence in a compact online
associative state read as low-rank attention corrections~\cite{lei2026deltamem}.
Such states are efficient and tightly coupled to generation, but they also mix
heterogeneous evidence in a shared representation. LifeFuse-Mem addresses the
resulting write interference by adding lifecycle-aware subspace isolation and
phase-aware readout to an existing parametric memory.

\subsection{Continual Learning and Catastrophic Forgetting}
Continual learning studies retention under sequential updates. Common
solutions include replay and episodic constraints~\cite{yang2025rethinklongcontext,wang2025workflowmemory},
importance-based regularization~\cite{liao2026sparta,li2025ticlm}, 
parameter isolation~\cite{hu2025hiagent,liao2026sparta}, and
subspace or gradient-projection methods~\cite{liao2026sparta,feng2025recurrentkif}; recent LLM work
adapts related ideas through orthogonal low-rank adaptation~\cite{liao2026sparta,yang2025rethinklongcontext}.
Surveys provide broader taxonomies~\cite{yang2025foundationcl}. Temporary
overwrite is related but has a different target behavior: the later value may
be locally valid without being a permanent replacement. LifeFuse-Mem therefore
uses isolation and routing to distinguish information lifecycles inside one
online memory state, rather than to retain old task behavior across discrete
tasks.

\subsection{Long-Term Memory Evaluation}
Long-term memory benchmarks test whether models can use extended histories,
including conversational and agent-oriented settings such as
LoCoMo~\cite{maharana2024locomo} and LongMemEval~\cite{wu2025longmemeval},
multitask long-context suites~\cite{bai2025longbenchv2,hu2026memoryagentbench,yen2025helmet}, bilingual
long-context understanding~\cite{bai2025longbenchv2}, evaluation beyond 100K
tokens~\cite{bai2025longbenchv2,wu2025lifbench}, and controlled retrieval or aggregation
stress tests~\cite{wu2025lifbench,tan2025membench,zhang2025memsim,modarressi2025nolima}. These benchmarks are
valuable for general memory ability, but they typically do not separate
permanent facts from temporary overwrites or condition retention on successful
initial acquisition. Our Hard Attribution Anti-Overwrite benchmark complements
them with an explicit Phase-A/Phase-B conflict structure and
acquisition-controlled overwrite metrics.

\section{Preliminaries and Problem Formulation}
\label{sec:problem}

\subsection{Memory-Conditioned Response Generation}

We consider a long-running language agent that receives a query $x$ and
generates a response $y$ using both the parametric model and accumulated
memory. The relevant memory is not homogeneous. Permanent facts
$P=\{(k_i,v_i^p)\}_{i=1}^{n}$ encode stable user facts or durable task
preferences whose values should remain valid across future interactions.
Temporary facts $T=\{(k_j,v_j^t)\}_{j=1}^{m}$ encode local assumptions,
task-specific conditions, or counterfactual instructions whose values should
apply only within the current context.

A response therefore depends on both semantic relevance and information
lifecycle. A permanent query should recover $v_i^p$ even after later local
context is written, whereas a temporary query should follow the active
temporary value $v_j^t$. The hard case occurs when temporary and permanent
facts share the same semantic attribute but assign conflicting values. The
model must use the temporary value for the local response without making it
authoritative for future permanent queries.

\subsection{Lifecycle Conflict in Compact Online Memory}

A broad class of neural memory methods compresses interaction history into a
compact online state. At injected layer $l$, we write this state as
$S_t^{(l)} \in \mathbb{R}^{r \times r}$. Given hidden state $h_t$, the memory
module produces
\[
q_t^m=\mathrm{norm}(W_q^m h_t),
\]
\[
k_t^m=\mathrm{norm}(W_k^m h_t),\quad v_t^m=W_v^m h_t.
\]
The state is read as $z_t=S_{t-1}q_t^m$ and injected back into the model
computation. A token-level write gate
$\beta_t=\sigma(W_\beta h_t+b_\beta)$ and keep coefficient
$\lambda_t=1-\beta_t$ define a common delta-style update:
\[
S_t=\mathrm{diag}(\lambda_t)S_{t-1}
 + \mathrm{diag}(\beta_t)(v_t^m-S_{t-1}k_t^m)(k_t^m)^\top .
\]
This abstraction covers compact neural memory adapters that read and write a
shared low-rank state, including $\delta$-Mem.

The lifecycle conflict appears when the same state stores both lifecycles.
After Phase A, state $S_A$ should contain permanent facts; after Phase B, state
$S_{AB}$ includes temporary overwrites. If both phases use the same rows and
write dynamics, Phase B can erase or re-rank evidence for the Phase-A value. We
call this \emph{temporary overwrite}: after writing a temporary value, the
memory state answers a later permanent query with that temporary value.

This failure differs from ordinary acquisition failure, where the permanent
value may never have been stored. We therefore evaluate retention only on facts
acquired after Phase A. Let $A_i=1$ if the model answers the $i$-th permanent
query correctly after Phase A, and let $R_i=1$ if it still returns the
permanent value after Phase B. The acquisition-controlled retention is
\[
\mathrm{Ret}_{acq} =
\frac{\sum_i \mathbf{1}[A_i=1 \wedge R_i=1]}
{\sum_i \mathbf{1}[A_i=1]}.
\]
The overwrite rate is computed on the same acquired subset, counting answers
equal to the conflicting temporary value. This protocol isolates the lifecycle
question: once a permanent fact has been acquired, does a temporary write
incorrectly become authoritative?

\section{LifeFuse-Mem}
\label{sec:lifefuse}

\subsection{Framework Overview}

LifeFuse-Mem adds a lifecycle boundary to compact neural memory. The framework
has three parts, shown in Figure~\ref{fig:method-overview}. First, it learns a
plastic/stable subspace partition inside each online memory state. Second, a
supervised permanent router controls how strongly write tokens can update the
stable rows. Third, phase-aware protected-state readout is used only in the controlled overwrite-attribution protocol, where phase boundaries and query lifecycles are known in advance, to evaluate whether preserved evidence can be recovered after conflicting temporary writes. Thus, lifecycle control is applied at writing, training, and readout,
without returning answers from an external key--value table.

\subsection{Lifecycle-Aware Subspace Partition}

A fixed split of raw rank dimensions is arbitrary, so LifeFuse-Mem learns an
approximately orthogonal basis for the memory coordinates in each injected
attention module. Let $R_l\in\mathbb{R}^{r\times r}$ be a learned square
matrix and let $Q_l=\operatorname{orth}(R_l)$ be the orthogonalized basis used
in the forward pass, with $Q_l^\top Q_l=I$ up to numerical precision. The
memory read, key, and value vectors are rotated into this basis before
subspace-specific control:
\[
\tilde q_t=Q_l^\top q_t^m,\quad
\tilde k_t=Q_l^\top k_t^m,\quad
\tilde v_t=Q_l^\top v_t^m .
\]
The first $r_A$ coordinates define the plastic subspace $A$, and the remaining
$r_B=r-r_A$ coordinates define the stable subspace $B$. In the experiments,
$r=8$ and the split is $r_A=r_B=4$. The raw rotation parameter is regularized by
\[
\mathcal{L}_{orth}=\frac{1}{Lr^2}\sum_{l=1}^{L}\|R_l^\top R_l-I\|_F^2,
\]
where $L$ is the number of injected memory modules. The $1/r^2$ factor matches
the elementwise mean used in the implementation and does not change the
minimizer. This regularizer keeps the subspaces separable while allowing the
model to learn a coordinate system more suitable than raw rank indices.

The underlying state is the same compact associative state used by $\delta$-Mem.
For a single memory module, $S_t\in\mathbb{R}^{r\times r}$ is read and updated
online. Before token $t$ writes, the memory read is
\[
z_t=S_{t-1}\tilde q_t .
\]
With row-wise keep, erase, and write coefficients
$\kappa_t,\epsilon_t,\omega_t\in\mathbb{R}^{r}$, the affine update is
\[
S_t=\kappa_t\odot_{\mathrm{row}} S_{t-1}
-\epsilon_t\odot_{\mathrm{row}}\left[(S_{t-1}\tilde k_t)\tilde k_t^\top\right]
+\omega_t\odot_{\mathrm{row}}\left[\tilde v_t\tilde k_t^\top\right].
\]
The read vector is rotated back before it is projected into the active
attention heads.

\subsection{Permanent Routing and Subspace Writing}

For each write token, LifeFuse-Mem predicts a permanent-route weight
\[
\rho_t=\sigma(w_r^\top h_t+b_r).
\]
In the implemented training data, the route target comes from the episode
metadata. If the target lifecycle is permanent, valid write tokens in that
example receive $y_i=1$; if it is temporary, they receive $y_i=0$; examples
without a lifecycle type are masked. The route supervision loss is
\[
\mathcal{L}_{route} =
-\frac{1}{|\Omega|}\sum_{(i,t)\in\Omega}
\left[y_i\log\rho_{i,t}+(1-y_i)\log(1-\rho_{i,t})\right],
\]
where $\Omega$ is the set of valid routed write tokens.

The router affects writing through row scaling. Plastic rows remain writable
for all tokens. Stable rows use an interpolation between a temporary scale
$a_{\mathrm{tmp}}$ and a permanent scale $a_{\mathrm{perm}}$:
\[
a_{t,j}=
\begin{cases}
1, & j\le r_A,\\
a_{\mathrm{tmp}}+\rho_t(a_{\mathrm{perm}}-a_{\mathrm{tmp}}), & j>r_A .
\end{cases}
\]
For the main LifeFuse-Mem runs, the implementation uses the
\texttt{permanent\_boost} mode with $a_{\mathrm{tmp}}=0$ and
$a_{\mathrm{perm}}=1$. Under the standard $\delta$-Mem update this row scale turns
the base coefficients into
$\kappa_{t,j}=1-a_{t,j}+a_{t,j}\lambda_{t,j}$ and
$\epsilon_{t,j}=\omega_{t,j}=a_{t,j}\beta_{t,j}$. Thus tokens with high
$\rho_t$ can update both plastic and stable rows, while tokens with low
$\rho_t$ primarily update the plastic rows because stable-row erase and write
coefficients are suppressed. The implementation also applies subspace key isolation:
during ordinary online writing, stable key coordinates are attenuated
($0.5$ in the reproduced runs), so writes prefer plastic coordinates on the key
axis as well. This design does not hard-code a separate memory module for each
lifetime; instead, route labels, row scaling, key isolation, and downstream
ranking losses encourage lifecycle-aware separation under explicit route
supervision.

\subsection{Episodic Learning Objective}

LifeFuse-Mem is trained with an episodic write-then-read objective. The model
first consumes write messages and updates memory to $S_{\mathrm{write}}$. It
then answers the final assistant target with further writing disabled, forcing
the answer to depend on the stored state rather than updating memory during the
read. The full objective is
\[
\begin{array}{rl}
\mathcal{L}= & \mathcal{L}_{CE}
+\lambda_{route}\mathcal{L}_{route}
+\lambda_{orth}\mathcal{L}_{orth} \\
&+\lambda_{hc}(u)\mathcal{L}_{hc}
+\lambda_{ret}\mathcal{L}_{ret}
+\lambda_{sp}\mathcal{L}_{sp}.
\end{array}
\]
where $u$ is training progress. The cross-entropy term trains the adapter state
to support the target answer after context writing.

The hard candidate loss improves candidate discrimination. Given the correct
answer $y^+$ and a hard negative $y^-$, we compute their average answer
cross-entropies $CE^+$ and $CE^-$. The objective encourages the negative
candidate to have loss at least $m_{hc}$ larger than the correct candidate:
\[
\mathcal{L}_{hc}=\tau_{hc}\,
\mathrm{softplus}\left(\frac{m_{hc}-(CE^- - CE^+)}{\tau_{hc}}\right).
\]
This loss is ramped in after early answer learning has stabilized. For
permanent examples, a retention-ranking loss with the same form uses its own
margin $m_{ret}$ and temperature $\tau_{ret}$, giving extra pressure for the
permanent value to outrank the temporary or rejected candidate. Finally, a
sparsity penalty keeps the average write gate near a small target $\beta_0$
($0.05$ in the reproduced runs), discouraging unnecessary memory updates on
every token.

\subsection{Protected State Fusion and Lifecycle Readout}

This component is specific to the controlled overwrite-attribution protocol and should not be interpreted as autonomous lifecycle discovery. In that setting, LifeFuse-Mem checkpoints the memory
state after Phase A as $S_A$ and also keeps the post-overwrite state
$S_{AB}$. It then constructs a stable-row protected state
$S_{\mathrm{stable}}$ by taking plastic rows from $S_{AB}$ and stable rows
from $S_A$:
\[
S_{\mathrm{stable}}[..., :r_A, :] = S_{AB}[..., :r_A, :],
\]
\[
S_{\mathrm{stable}}[..., r_A:, :] = S_A[..., r_A:, :] .
\]
Thus, $S_{\mathrm{stable}}$ preserves the part of memory intended to carry
long-lived facts while retaining recent plastic context.

For a permanent query, the model scores each candidate $y\in\mathcal{C}$ under
both $S_A$ and $S_{\mathrm{stable}}$. The candidate score is the
length-normalized token log probability
\[
s_S(y)=\frac{1}{|y|}\sum_{\tau=1}^{|y|}
\log p_\theta(y_\tau\mid x,y_{<\tau},S).
\]
The two component scores are
\[
s_A(y)=s_{S_A}(y),
\]
\[
s_{\mathrm{stable}}(y)=s_{S_{\mathrm{stable}}}(y).
\]
The final permanent-query score is
\[
s_{\mathrm{fuse}}(y)=\alpha s_A(y)+(1-\alpha)s_{\mathrm{stable}}(y),
\]
and the selected answer is
\[
\hat y=\arg\max_{y\in\mathcal{C}}s_{\mathrm{fuse}}(y).
\]
For a temporary query, LifeFuse-Mem does not protect or fuse the state; it
scores candidates with the full post-overwrite state $S_{AB}$. The readout
therefore matches the intended lifecycle of the query: permanent queries
consult protected evidence, whereas temporary queries follow the local
overwrite.

\section{Experiments}

\subsection{Experimental Setup}

\paragraph{Experimental Design}
We evaluate LifeFuse-Mem from three angles. First, we use a targeted
temporary-overwrite benchmark to test whether permanent facts remain
recoverable after conflicting temporary writes. Second, we evaluate on public
long-memory benchmarks to check whether the proposed lifecycle boundary
preserves general memory behavior outside the controlled overwrite setting.
Third, we ablate the main components on the targeted benchmark to identify the
source of the retention effect.

\paragraph{Evaluation Protocol}
We report results on Qwen3-4B and SmolLM3-3B. The controlled temporary-
overwrite benchmark is our main diagnostic experiment, while LoCoMo and
MemoryAgentBench test compatibility with standard long-memory evaluation. All
comparisons are made within the same backbone and use the same data split,
prompting format, scoring protocol, and memory rank. Trained adapters share the
same 3,243-sample training set and optimization setup, with details provided in
the Appendix file. LifeFuse-Mem additionally uses episode-level lifecycle
metadata for route supervision, so we include a no-route-supervision ablation
to isolate this signal.

\paragraph{Shared Memory Configuration}
All comparison experiments use the same rank-8 LoRA configuration for the
corresponding backbone. LifeFuse-Mem changes only the lifecycle-aware components: it
splits the rank into four plastic and four stable dimensions, learns an
orthogonal basis for this partition, enables token-level permanent routing, and
uses the losses defined in Section~\ref{sec:lifefuse}. The basis is
orthonormalized by QR decomposition at each forward pass; the orthogonal
regularizer acts on the underlying parameter and is not the source of exact
orthogonality. On Hard Attribution Anti-Overwrite, where Phase-A/Phase-B
boundaries and query lifecycles are known, permanent queries use phase-aware
fusion readout: LifeFuse-Mem fuses the Phase-A state with a
stable-row-protected post-overwrite state, where plastic rows come from the
post-overwrite state and stable rows are restored from Phase A. Temporary
queries use the full post-overwrite state. The main fusion weight is
$\alpha=0.70$; the Appendix file provides the detailed sensitivity ablation.
On LoCoMo and MemoryAgentBench, which do not provide overwrite phases or query
lifecycle labels, LifeFuse-Mem is evaluated in standard online-memory mode:
the model consumes the benchmark history with normal memory writes and answers
from the resulting memory state, without Phase-A checkpointing,
stable-row restoration, or query-type-specific fusion.

\begin{table}[t]
\centering
\small
\caption{Results on Hard Attribution Anti-Overwrite. Scores are percentages. Ret. and Ovr. are acquisition-controlled metrics computed only over conflict-permanent facts answered correctly after Phase A; under this choice protocol, they are complementary. Temp. is an auxiliary sanity check on temporary-query behavior. Higher is better except Ovr.}
\label{tab:hard-attribution}
\begin{tabular}{lrrrr}
\toprule
Model & \multicolumn{3}{c}{Permanent Query} & Temp. (aux.) \\
\cmidrule(lr){2-4}\cmidrule(l){5-5}
 & Perm. & Ret. & Ovr. $\downarrow$ & Temp. \\
\midrule
\textbf{Qwen3-4B} & & & & \\
\quad + LoRA-Mem & 19.65 & 61.03 & 38.97 & 20.13 \\
\quad + Titans & 19.65 & 58.02 & 41.98 & 20.23 \\
\quad + $\delta$-Mem & 19.32 & 57.72 & 42.28 & \textbf{20.27} \\
\quad + LifeFuse-Mem & \textbf{20.12} & \textbf{63.71} & \textbf{36.29} & 20.10 \\
\midrule
\textbf{SmolLM3-3B} & & & & \\
\quad + LoRA-Mem & 19.68 & 61.14 & 38.86 & 19.95 \\
\quad + Titans & 19.82 & 61.71 & 38.29 & 19.88 \\
\quad + $\delta$-Mem & \textbf{19.95} & 60.27 & 39.73 & 20.18 \\
\quad + LifeFuse-Mem & 19.65 & \textbf{69.39} & \textbf{30.61} & \textbf{20.35} \\
\bottomrule
\end{tabular}
\end{table}

\begin{table*}[t!]
\centering
\small
\caption{Results on public memory benchmarks. Scores are percentages. MAB denotes MemoryAgentBench. Bold values indicate the best method under the same backbone and benchmark category.}
\label{tab:public-benchmarks}
\resizebox{\textwidth}{!}{%
\begin{tabular}{lrrrrrrrrrr}
\toprule
Model & \multicolumn{5}{c}{MemoryAgentBench} & \multicolumn{5}{c}{LoCoMo} \\
\cmidrule(lr){2-6}\cmidrule(l){7-11}
 & Avg. & AR & TTL & LRU & SF & Avg. & Multi & Temp. & Open & Single \\
\midrule
\textbf{Qwen3-4B} & & & & & & & & & & \\
\quad + LoRA-Mem & 31.46 & 38.80 & 23.93 & 46.37 & 16.50 & 39.95 & 28.57 & 28.05 & 15.19 & 51.14 \\
\quad + Titans & 30.56 & 40.00 & 18.86 & 46.25 & 17.13 & 34.33 & 26.60 & 21.64 & 13.26 & 44.17 \\
\quad + $\delta$-Mem & 31.83 & \textbf{40.20} & 21.00 & \textbf{46.52} & \textbf{17.25} & 41.24 & 29.78 & 29.57 & 14.57 & 52.57 \\
\quad + LifeFuse-Mem & \textbf{32.24} & \textbf{40.20} & \textbf{24.00} & 45.30 & 16.75 & \textbf{42.41} & \textbf{33.62} & \textbf{30.71} & \textbf{16.17} & \textbf{52.81} \\
\midrule

\textbf{SmolLM3-3B} & & & & & & & & & & \\
\quad + LoRA-Mem & 19.85 & 25.25 & 8.57 & \textbf{37.20} & 12.50 & 30.79 & 21.69 & 23.40 & 14.35 & 38.53 \\
\quad + Titans & 19.39 & \textbf{27.85} & 4.79 & 37.17 & 7.75 & 23.82 & 20.82 & 16.95 & 12.82 & 28.70 \\
\quad + $\delta$-Mem & 20.13 & 25.55 & 8.50 & 37.12 & \textbf{13.13} & 30.65 & 22.38 & 23.19 & 14.99 & 38.06 \\
\quad + LifeFuse-Mem & \textbf{20.21} & 25.85 & \textbf{9.05} & 36.55 & 12.37 & \textbf{38.30} & \textbf{28.47} & \textbf{27.08} & \textbf{18.50} & \textbf{48.14} \\
\bottomrule
\end{tabular}%
}
\end{table*}

\subsection{Hard Attribution Anti-Overwrite}
\label{sec:temporary-overwrite}
\paragraph{Benchmark and Metrics}
We construct a 1,000-episode Hard Attribution Anti-Overwrite benchmark to
measure the target failure mode directly. Each episode writes permanent facts
in Phase A and then introduces conflicting temporary facts in Phase B. Permanent
queries should recover the Phase-A value, whereas temporary queries should
follow the Phase-B value. This separates stable-fact retention from local
adaptation to temporary context.

The benchmark is intentionally difficult. Queries use natural-language aliases
rather than canonical keys, some answers require cross-sentence attribution,
both phases contain noise facts, conflicts may be indirect, and queries are
scored with hard-candidate choice sets. Each episode contains two entities, four
protected facts per entity, six conflict keys, four overwrite rounds, and noise
facts in both phases.

Table~\ref{tab:hard-attribution} reports four metrics. \textbf{Perm.} is final
accuracy on all permanent queries after Phase B, using the Phase-A value as the
target. Because this metric also depends on initial acquisition, we additionally
report acquisition-controlled metrics over conflict-permanent facts answered
correctly after Phase A. \textbf{Ret.} is the fraction of this acquired subset
still answered with the permanent value after Phase B. \textbf{Ovr.} is the
fraction answered with the conflicting temporary value after Phase B; lower is
better. In the released hard-candidate choice evaluation, each acquired
conflict-permanent query is ranked over the Phase-A permanent answer and four
temporary overwrite candidates. Under this protocol, Ret. and Ovr. are
complementary; this complementarity is not a general property of the metrics
under open-generation or mixed-distractor evaluations. \textbf{Temp.} is final
accuracy on temporary queries, using the Phase-B value
as the target. We report Temp. as an auxiliary sanity check on Phase-B
behavior, not as the main criterion for overwrite. Ret. does not measure how
many facts are acquired initially, so it should be read together with Perm.

\paragraph{Results}
Table~\ref{tab:hard-attribution} shows the clearest gain on the
acquisition-controlled metrics. Compared with $\delta$-Mem, LifeFuse-Mem
raises Ret. from 57.72\% to 63.71\% on Qwen3-4B and from 60.27\% to 69.39\% on
SmolLM3-3B. Ovr. decreases in parallel, from 42.28\% to 36.29\% on Qwen3-4B and
from 39.73\% to 30.61\% on SmolLM3-3B. Thus, among facts that were already
acquired after Phase A, LifeFuse-Mem is less likely to answer with the
conflicting temporary value after Phase B.



End-to-end permanent-query accuracy changes only marginally: it increases on Qwen3-4B, from 19.32\% to 20.12\%, but changes from 19.95\% to 19.65\% on SmolLM3-3B. This indicates that LifeFuse-Mem primarily improves overwrite resistance conditional on successful acquisition, rather than substantially improving the overall probability of answering permanent queries correctly end to end. Temp. remains similar across methods, which is consistent with the retention gain not coming from simply suppressing Phase-B information. Accordingly, Ret. and Ovr. should be interpreted as diagnostic measures of overwrite behavior, not as substitutes for end-to-end task accuracy.

\begin{table}[htbp]
\centering
\small
\caption{Component ablations on Hard Attribution Anti-Overwrite. Scores are percentages. Ret. and Ovr. are acquisition-controlled and complementary under the released hard-candidate choice protocol; Temp. is auxiliary; lower Ovr. is better.}
\label{tab:ablations}
\begin{tabular}{lrrrr}
\toprule
Model & \multicolumn{3}{c}{Permanent Query} & Temp. (aux.) \\
\cmidrule(lr){2-4}\cmidrule(l){5-5}
 & Perm. & Ret. & Ovr. $\downarrow$ & Temp. \\
\midrule
\textbf{Qwen3-4B} & & & & \\
\quad + LifeFuse-Mem & \textbf{20.12} & 63.71 & 36.29 & 20.10 \\
\quad w/o route sup. & 19.88 & 58.68 & 41.32 & 19.98 \\
\quad w/o stable rows & 19.93 & 62.79 & 37.21 & 20.10 \\
\quad w/o hard cand. & 19.28 & \textbf{64.09} & \textbf{35.91} & \textbf{20.55} \\
\quad w/o ret. loss & 20.00 & 62.37 & 37.63 & 19.15 \\
\midrule
\textbf{SmolLM3-3B} & & & & \\
\quad + LifeFuse-Mem & 19.65 & \textbf{69.39} & \textbf{30.61} & 20.35 \\
\quad w/o route sup. & 19.62 & 59.45 & 40.55 & \textbf{20.38} \\
\quad w/o stable rows & 19.72 & 68.47 & 31.53 & 20.22 \\
\quad w/o hard cand. & 19.98 & 69.04 & 30.96 & 20.35 \\
\quad w/o ret. loss & \textbf{20.10} & 65.74 & 34.26 & 19.98 \\
\bottomrule
\end{tabular}
\end{table}

\subsection{Long-Memory Benchmarks}
\label{sec:public-benchmarks}

\paragraph{Benchmark and Metrics}
We next evaluate whether the lifecycle boundary remains compatible with public
long-memory tasks. This is a compatibility check rather than a direct
anti-overwrite test: a method could protect old facts on the targeted benchmark
while degrading standard memory behavior. We use LoCoMo and MemoryAgentBench.
LoCoMo is evaluated with full-history replay on categories 1--4, covering 1,540
questions, and reports overall, multi-hop, temporal, open-domain, and
single-hop scores. MemoryAgentBench contains 3,671 questions from 14 datasets,
grouped into Accurate Retrieval (AR), Test-time Learning (TTL), Long Range
Understanding (LRU), and Selective Forgetting (SF). For MemoryAgentBench, we
use a unified prompt template, temperature 0.4, top-$p$ 0.9, top-$k$ 10, maximum
generation length 4,096, maximum context length 120,000 characters, and
evaluation batch size 16. Its overall score is the sample-weighted category
average. Since these public benchmarks do not annotate information lifecycles
or controlled overwrite phases, no protected-state fusion is used in this
evaluation. The learned router and subspace partition remain active as part of
the memory adapter, but no benchmark-specific lifecycle labels are supplied at
inference time.

\paragraph{Results}
Table~\ref{tab:public-benchmarks} reports compatibility results on public
long-memory benchmarks. LifeFuse-Mem remains broadly competitive: relative to
$\delta$-Mem on Qwen3-4B, LoCoMo improves from 41.24\% to 42.41\% and
MemoryAgentBench increases slightly from
31.83\% to 32.24\%. Category-level results vary: TTL improves from 21.00\% to
24.00\%, AR ties at 40.20\%, while LRU and SF decrease from 46.52\% to 45.30\% and
from 17.25\% to 16.75\%, respectively.

Relative to $\delta$-Mem on SmolLM3-3B, LifeFuse-Mem also improves LoCoMo
overall score, from 30.65\% to 38.30\%, with gains in all LoCoMo categories. On
MemoryAgentBench, the overall score is again close to $\delta$-Mem, increasing
slightly from 20.13\% to 20.21\%. AR and TTL improve from 25.55\% to 25.85\% and
from 8.50\% to 9.05\%, while LRU and SF decrease from 37.12\% to 36.55\% and from
13.13\% to 12.37\%, respectively. This suggests that the lifecycle boundary
improves some public-benchmark categories while reducing others, so its benefits
are not uniform across all MemoryAgentBench task types.

Neither LoCoMo nor MemoryAgentBench instantiates the Phase-A/Phase-B overwrite
structure or separates permanent from temporary queries. They should therefore
be interpreted as compatibility checks for running the trained lifecycle-aware
memory adapter under ordinary long-memory inference, not as direct evidence for
the phase-aware anti-overwrite readout. The targeted benchmark provides the
direct anti-overwrite evidence; the public benchmarks show how the same trained
memory architecture affects general long-memory behavior when explicit
lifecycle boundaries are absent.

\subsection{Ablation Studies}
\label{sec:ablations}

\paragraph{Component Ablations}
We ablate four components on the 1,000-episode Hard Attribution benchmark:
route supervision, stable rows, hard candidate loss, and permanent
retention loss. Each ablation removes one component and keeps the remaining
training and inference settings unchanged. We report the same metrics as in
Section~\ref{sec:temporary-overwrite} to measure retention and overwrite under
the same acquisition-controlled protocol. The main fusion weight is
$\alpha=0.70$; the Appendix file reports the detailed weight-sensitivity
ablation.

\paragraph{Ablation Results}
Table~\ref{tab:ablations} identifies route supervision as the most important
component. Removing it decreases retention from 63.71\% to 58.68\% on Qwen3-4B and
from 69.39\% to 59.45\% on SmolLM3-3B; overwrite rises to 41.32\% and 40.55\%,
respectively. This indicates that rank partitioning alone is insufficient:
token-level permanent routing provides the signal that makes the subspaces
lifecycle-aware.

The remaining components contribute in more model-dependent ways. Removing
stable rows or permanent retention loss consistently weakens
acquisition-controlled retention relative to the full model on both backbones.
Removing hard candidate loss is not uniformly beneficial or harmful. On
Qwen3-4B, Ret. and Ovr. improve slightly, but end-to-end Perm. drops from
20.12\% to 19.28\%, indicating a trade-off between the acquired subset and final
permanent-query accuracy. On SmolLM3-3B, its effect is small relative to route
supervision and retention loss. Overall, the ablations identify route
supervision as the most reliable contributor to acquisition-controlled
retention. Stable rows and the retention loss provide additional but
smaller gains, while the hard-candidate loss mainly affects candidate
calibration and does not uniformly improve Ret. or Ovr. Thus, the evidence
supports the role of route-conditioned isolation, but does not imply that every
auxiliary objective is necessary for every metric.

\section{Conclusion}
This work studies temporary overwrite as a lifecycle mismatch failure mode in compact online memory and introduces LifeFuse-Mem as a targeted mitigation framework. By combining lifecycle-aware routing, subspace isolation, and protected readout, LifeFuse-Mem reduces acquisition-conditioned overwrite on the controlled Hard Attribution Anti-Overwrite benchmark. Results on LoCoMo and MemoryAgentBench indicate that the trained adapter remains compatible with standard long-memory evaluation when protected-state fusion is not available, although the gains vary across backbones and task categories. Ablation studies identify route supervision as the most reliable source of the retention gains, with stable rows, protected fusion, and auxiliary losses contributing more selectively. Overall, the results support explicit lifecycle modeling as a useful diagnostic and architectural principle for long-term memory, while leaving lifecycle inference from unlabeled interaction histories to future work.



\bibliography{main}
\bibliographystyle{plainnat}


\end{document}